\documentclass[twoside,11pt]{article}
\usepackage{jmlr2e}
\usepackage{tcolorbox}
\usepackage{color-edits}
\usepackage[scaled=0.8]{DejaVuSansMono}
\usepackage{enumitem}
\usepackage{booktabs}
\usepackage{float}
\usepackage{listings}
\usepackage{xcolor}
\usepackage{amsmath}

\usepackage{lastpage}

\usepackage{xcolor}

\definecolor{dark-blue}{rgb}{0.15,0.15,0.4}
\definecolor{codegreen}{rgb}{0,0.6,0}
\definecolor{codegray}{rgb}{0.5,0.5,0.5}
\definecolor{codepurple}{rgb}{0.58,0,0.82}
\definecolor{backcolour}{rgb}{0.95,0.95,0.92}

\lstdefinestyle{mystyle}{
    backgroundcolor=\color{backcolour},   
    commentstyle=\color{codegreen},
    keywordstyle=\color{magenta},
    numberstyle=\tiny\color{codegray},
    stringstyle=\color{codepurple},
    basicstyle=\ttfamily\scriptsize\color{blue!30!black},    emph={int,char,double,float,unsigned,void,bool},
    emphstyle={\color{blue}},
    morekeywords={>,<,.,;,-,!,=,~},
    otherkeywords={>,<,.,;,-,!,=,~},
    breakatwhitespace=false,         
    breaklines=false,             
    language=python,
    captionpos=b,                    
    keepspaces=true,                   
    showspaces=false,                
    showstringspaces=false,
    showtabs=false,                  
    tabsize=4
}
\hypersetup{ hidelinks }
\firstpageno{1}
\begin{document}
\emergencystretch 3em
\title{UQLM: A Python Package for Uncertainty Quantification in Large Language Models}

\jmlrheading{27}{2026}{1-\pageref{LastPage}}{7/25; Revised
11/25}{1/26}{25-1557}{Dylan Bouchard, Mohit Singh Chauhan, David Skarbrevik, Ho-Kyeong Ra, Viren Bajaj, and Zeya Ahmad}
\ShortHeadings{UQLM: Uncertainty Quantification for Language Models}{Bouchard, Chauhan, Skarbrevik, Ra, Bajaj, and Ahmad}

\author{\name Dylan Bouchard\textsuperscript{1} \email dylan.bouchard@cvshealth.com \\
        \name Mohit Singh Chauhan\textsuperscript{1} \email mohitsingh.chauhan@cvshealth.com \\
        \name David Skarbrevik\textsuperscript{1} \email david.skarbrevik@cvshealth.com \\
        \name Ho-Kyeong Ra\textsuperscript{1} \email hokyeong.ra@cvshealth.com \\
        \name Viren Bajaj\textsuperscript{1} \email bajajv@aetna.com \\
        \name Zeya Ahmad\textsuperscript{1} \email zeya.ahmad@cvshealth.com \\ 
        \addr
            \textsuperscript{1}CVS Health, Wellesley, MA
       }
\editor{Joaquin Vanschoren}
\maketitle

\begin{abstract}%
Hallucinations, defined as instances where Large Language Models (LLMs) generate false or misleading content, pose a significant challenge that impacts the safety and trust of downstream applications. We introduce \texttt{uqlm}, a Python package for LLM  hallucination detection using state-of-the-art uncertainty quantification (UQ) techniques. This toolkit offers a suite of UQ-based scorers that compute response-level confidence scores ranging from 0 to 1. This library provides an off-the-shelf solution for UQ-based hallucination detection that can be easily integrated to enhance the reliability of LLM outputs.
\end{abstract}


\section{Introduction}

Large language models (LLMs) have revolutionized the field of natural language processing, but their tendency to generate false or misleading content, known as hallucinations, significantly compromises safety and trust. LLM hallucinations are especially problematic because they often appear plausible, making them difficult to detect and posing serious risks in high-stakes domains such as healthcare, legal, and financial applications. As LLMs are increasingly deployed in real-world settings, monitoring and detecting hallucinations becomes crucial.

Traditional evaluation methods involve `grading' LLM responses by comparing model output to human-authored ground-truth texts, an approach offered by toolkits such as Evals \citep{Openai} and G-Eval \citep{liu-etal-2023-g}. While effective during pre-deployment testing, these methods are limited in practice since users typically lack access to ground-truth data at generation time. This shortcoming motivates the need for generation-time hallucination detection methods.

Existing solutions to this problem include source-comparison methods, internet-based grounding, and uncertainty quantification (UQ) methods. Toolkits that offer source-comparison scorers, such as Ragas \citep{es2023ragasautomatedevaluationretrieval}, Phoenix \citep{phoenix}, DeepEval \citep{Ip_deepeval_2025}, and others \citep{ hu2024refcheckerreferencebasedfinegrainedhallucination, uptrain, zha-etal-2023-alignscore, asai2023selfraglearningretrievegenerate}
evaluate the consistency between generated content and input prompts. However, these methods can mistakenly validate responses that merely mimic prompt phrasing without ensuring factual accuracy. Toolkits that leverage Internet searches for fact-checking, such as FacTool \citep{chern2023factool}, introduce delays and risk incorporating erroneous online information, failing to address the inherent uncertainty in model outputs. Lastly, although numerous UQ techniques have been proposed in the literature, their adoption in user-friendly, comprehensive toolkits remains limited. For example, while SelfCheckGPT \citep{manakul-etal-2023-selfcheckgpt} 
 incorporates some UQ scorers, its set of techniques is narrow and does not integrate generation with evaluation, thus creating barriers for practitioners outside specialized AI research environments.
 LangKit \citep{langkit} and NeMo Guardrails \citep{rebedea-etal-2023-nemo} also offer UQ scorers but are similarly narrow in scope. LM-Polygraph \citep{fadeeva-etal-2023-lm} provides a valuable collection of UQ-based approaches for LLMs and is built on the Hugging Face ecosystem, representing an important step toward more accessible UQ tools, though opportunities remain for enhancing accessibility for non-specialized practitioners.

We aim to bridge these gaps by introducing a comprehensive open-source Python package, \texttt{uqlm}, that democratizes advanced research in LLM uncertainty quantification.  UQLM (Uncertainty Quantification for Language Models) implements a diverse array of uncertainty estimation techniques to compute generation-time, response-level confidence scores and uniquely integrates generation and evaluation processes. This integrated approach allows users to generate and assess content simultaneously, without the need for ground-truth data or external knowledge sources, and with minimal engineering effort. This democratization of access empowers smaller teams, researchers, and developers to incorporate robust hallucination detection into their applications for safer and more reliable AI systems.


\section{Usage}
The \texttt{uqlm} library, available at \url{https://github.com/cvs-health/uqlm}, provides a collection of UQ-based scorers spanning four categories: black-box UQ, white-box UQ, LLM-as-a-Judge, and ensembles.\footnote{For a detailed overview of available scorers and associated experiment results, we refer the reader to this project's companion paper, \cite{bouchard2025uncertainty}.} The corresponding classes for these techniques are instantiated by passing a LangChain \texttt{BaseChatModel} (LLM) instance to the constructor.\footnote{UQLM is compatible with all \hyperlink{https://python.langchain.com/docs/integrations/chat/}{LangChain Chat Models}. More details are provided in Appendix \ref{sec:models}.} Each of these classes contains a \texttt{generate\_and\_score} method, which generates LLM responses to a user provided list of prompts and computes response-level confidence scores, which range from 0 to 1.


\subsection{Black-Box Uncertainty Quantification}
\begin{figure}[t]
    \centering
    \includegraphics[width=0.9\linewidth]{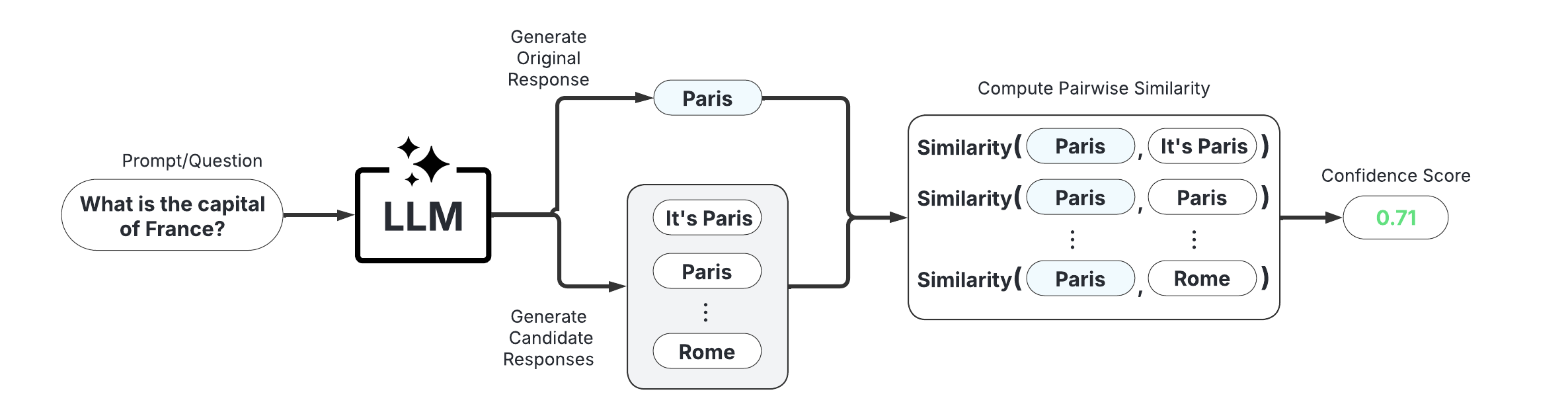}
    \caption{Illustration of a Black-Box Scorer Workflow}
    \label{fig:bb_graphic}
\end{figure}

Black-box uncertainty quantification exploits the stochastic nature of LLMs and measures the consistency of multiple responses to the same prompt. These consistency measurements can be conducted with various approaches, including discrete semantic entropy \citep{Farquhar2024}, number of semantic sets \citep{lin2024generatingconfidenceuncertaintyquantification}, non-contradiction probability \citep{chen-mueller-2024-quantifying}, entailment probability \citep{lin2024generatingconfidenceuncertaintyquantification}, BERTScore \citep{manakul-etal-2023-selfcheckgpt}, exact match rate \citep{cole-etal-2023-selectively}, and cosine similarity \citep{shorinwa2024surveyuncertaintyquantificationlarge}. Black-box UQ scorers are compatible with any LLM, but increase latency and generation costs. The \texttt{BlackBoxUQ} class uses the user-provided LLM to generate, for each prompt, an original response and additional candidate responses, then computes consistency scores using the specified scorers (see Figure \ref{fig:bb_graphic}). \footnote{Users may skip generation by calling the  \texttt{score} method with pre-generated responses. The workflow depicted in Figure \ref{fig:bb_graphic} does not apply to semantic entropy, which does not designate an `original response'.}  See below for a minimal \texttt{BlackBoxUQ} example.\footnote{\texttt{use\_best=True} selects the mode from the highest-probability semantic cluster \citep{Farquhar2024}.}

\begin{lstlisting}[language=Python]
from uqlm import BlackBoxUQ
bbuq = BlackBoxUQ(llm=llm, scorers=["exact_match", "noncontradiction"])
results = await bbuq.generate_and_score(prompts=prompts, num_responses=5, use_best=True)
\end{lstlisting}

\subsection{White-Box Uncertainty Quantification}
\begin{figure}[H]
    \centering
    \includegraphics[width=0.9\linewidth]{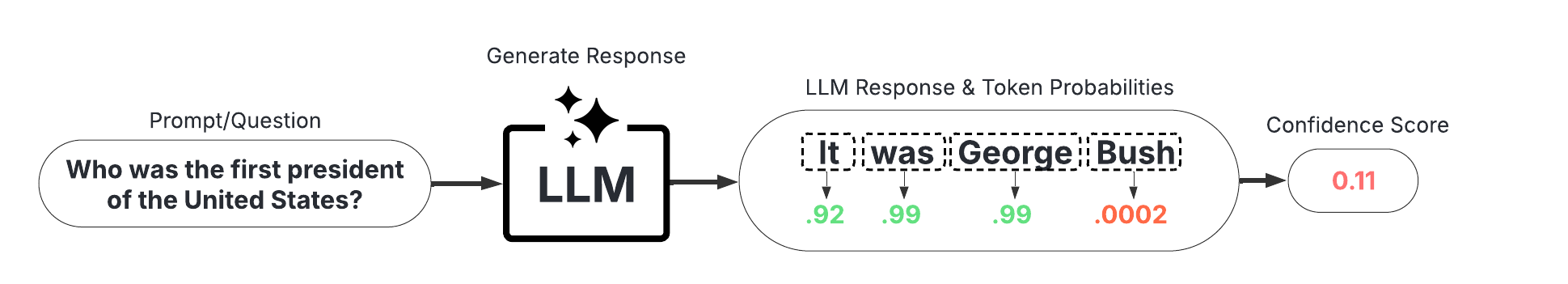}
    \caption{Illustration of a Single-Generation White-Box Scorer Workflow}
    \label{fig:wb_graphic}
\end{figure}

    White-box uncertainty quantification leverages token probabilities to compute uncertainty, as in Figure \ref{fig:wb_graphic}.    UQLM's \texttt{WhiteBoxUQ} class includes several categories of methods. Single-generation methods, which offer the advantage of no additional latency or generation costs, include minimum token probability \citep{manakul-etal-2023-selfcheckgpt}, length-normalized token probability \citep{malinin2021uncertaintyestimationautoregressivestructured}, sequence probability \citep{Vashurin_2025}, likelihood margin \citep{farr2024redctsystemsdesignmethodology}, mean top-K token entropy \citep{scalena2025eagerentropyawaregenerationadaptive}, and maximum top-K token entropy \citep{scalena2025eagerentropyawaregenerationadaptive}. Sampling-based methods include semantic entropy \citep{Farquhar2024}, semantic density \citep{qiu2024semanticdensityuncertaintyquantification}, Monte Carlo predictive entropy \citep{kuhn2023semanticuncertaintylinguisticinvariances}, and CoCoA \citep{vashurin2025uncertaintyquantificationllmsminimum}. Additionally, the class implements the $P(\text{True})$ method \citep{kadavath2022languagemodelsmostlyknow}, which requires one extra generation per response.  Note that WhiteBox UQ will only work with model APIs that expose token probabilities.  
    See below for example \texttt{WhiteBoxUQ} usage.

\begin{lstlisting}[language=Python]
from uqlm import WhiteBoxUQ
wbuq = WhiteBoxUQ(llm=llm, scorers=["min_probability"])
results = await wbuq.generate_and_score(prompts=prompts)
\end{lstlisting}

\subsection{LLM-as-a-Judge}
\begin{figure}[t]
    \centering
    \includegraphics[width=0.9\linewidth]{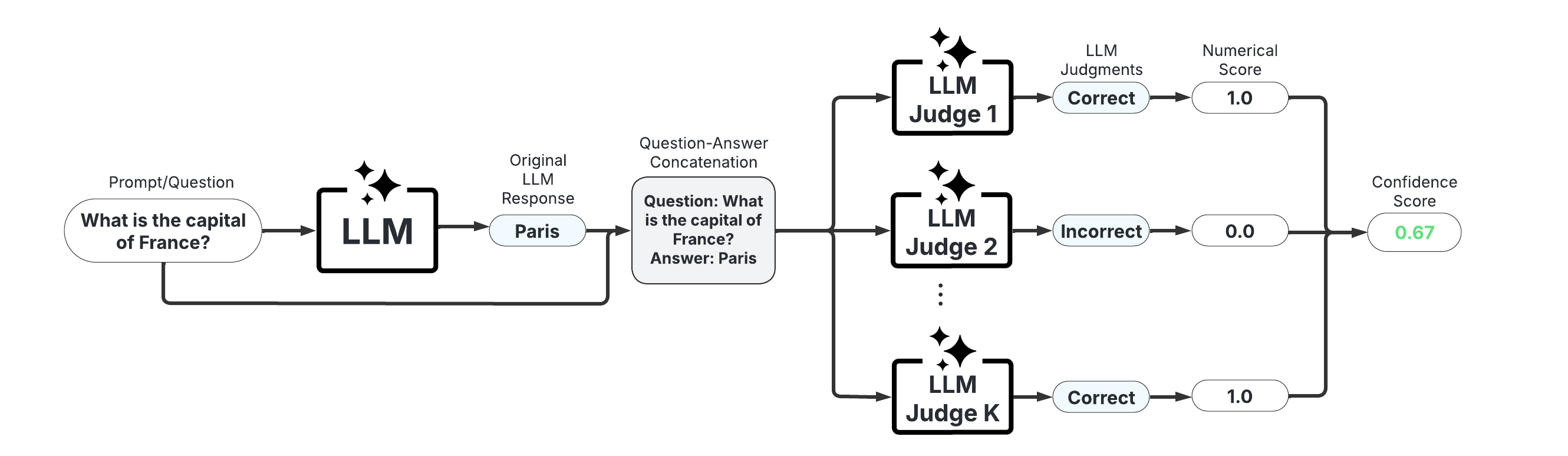}
    \caption{Illustration of LLM-as-a-Judge Workflow}
    \label{fig:judges_graphic}
\end{figure}

LLM-as-a-Judge uses an LLM to evaluate the correctness of a response to a particular question. To achieve this, a question-response concatenation is passed to one or more LLMs along with instructions to score the response's correctness using the \texttt{LLMPanel} class (see Figure \ref{fig:judges_graphic}). In the constructor, users pass a list of LLM objects to the \texttt{judges} argument and specify one of four scoring templates for each judge with the \texttt{scoring\_templates} argument.  These four scoring templates are as follows: binary ($\{ \text{incorrect}, \text{correct} \}$ as $\{ 0, 1 \}$), ternary ($\{ \text{incorrect}, \text{uncertain}, \text{correct}\}$ as $\{ 0, 0.5, 1 \}$), continuous (any value between 0 and 1), and a 5-point Likert scale ($0, 0.25,...,1$). Instructions are customized using the \texttt{additional\_context} argument. Implementing the \texttt{generate\_and\_score} method (minimal example below) returns the score from each judge along with aggregations such as average, minimum, and median.

\begin{lstlisting}[language=Python]
from uqlm import LLMPanel
panel = LLMPanel(llm=llm1, judges=[llm2, llm3], scoring_templates=["continuous", "likert"])
results = await panel.generate_and_score(prompts=prompts)
\end{lstlisting}

\subsection{Ensemble Approach}
Lastly, \texttt{uqlm} offers both tunable and off-the-shelf ensembles that leverage a weighted average of any combination of black-box UQ, white-box UQ, and LLM-as-a-Judge scorers. Similar to the aforementioned classes, \texttt{UQEnsemble} enables simultaneous generation and scoring with a \texttt{generate\_and\_score} method. Using the specified scorers, the ensemble score is computed as a weighted average of the individual confidence scores, where weights may be default weights, user-specified, or tuned (refer to Appendix \ref{appendix:tune} for tuning details). If no scorers are specified, the off-the-shelf implementation follows an ensemble of exact match, non-contradiction probability, and self-judge proposed by \cite{chen-mueller-2024-quantifying}.


\section{Conclusions}
This paper introduced \texttt{uqlm}, an open-source Python toolkit offering a collection of state-of-the-art uncertainty quantification techniques for LLM hallucination detection. We believe \texttt{uqlm} democratizes uncertainty quantification techniques from the literature, empowering practitioners to effectively detect hallucinations at generation time with minimal engineering effort. To get started with \texttt{uqlm}, we refer readers to the \href{https://github.com/cvs-health/uqlm}{Github repository} and \href{https://cvs-health.github.io/uqlm/latest/index.html}{Documentation site}.

\section*{Author Contributions}
Dylan Bouchard was the principal developer and researcher of the UQLM project, responsible for conceptualization, methodology, and software development of the UQLM library. Mohit Singh Chauhan helped lead research and software development efforts. David Skarbrevik, Ho-Kyeong Ra, Viren Bajaj, and Zeya Ahmad contributed to software development. 

\section*{Conflict of Interest}
The authors are employed and receive stock and equity from CVS Health® Corporation.

\acks{We wish to thank Piero Ferrante, Blake Aber, Matthew Churgin, Erik Widman,  Robert Enzmann, Xue (Crystal) Gu and Huiwen Hu for their helpful suggestions as well as Dimitris Tsapetis, Vipin Gyanchandani, Drew Ross, Jinesh Mehta, Namrata Walanj, and Joshua Mabry for their contributions to UQLM.}

\appendix

\section{Ensemble Tuning}
\label{appendix:tune}
In order to tune the ensemble weights prior to using the \texttt{generate\_and\_score} method, users must provide a list of prompts and corresponding ideal responses to serve as an `answer key'. The LLM's responses to the prompts are graded with a grader function that compares against the provided ideal responses. If a grader function is not provided by the user, the original LLM is used as an LLM-based grader to evaluate the correctness of the responses relative to the answer key. 

\begin{figure}[b]
    \centering
    \includegraphics[width=0.9\linewidth]{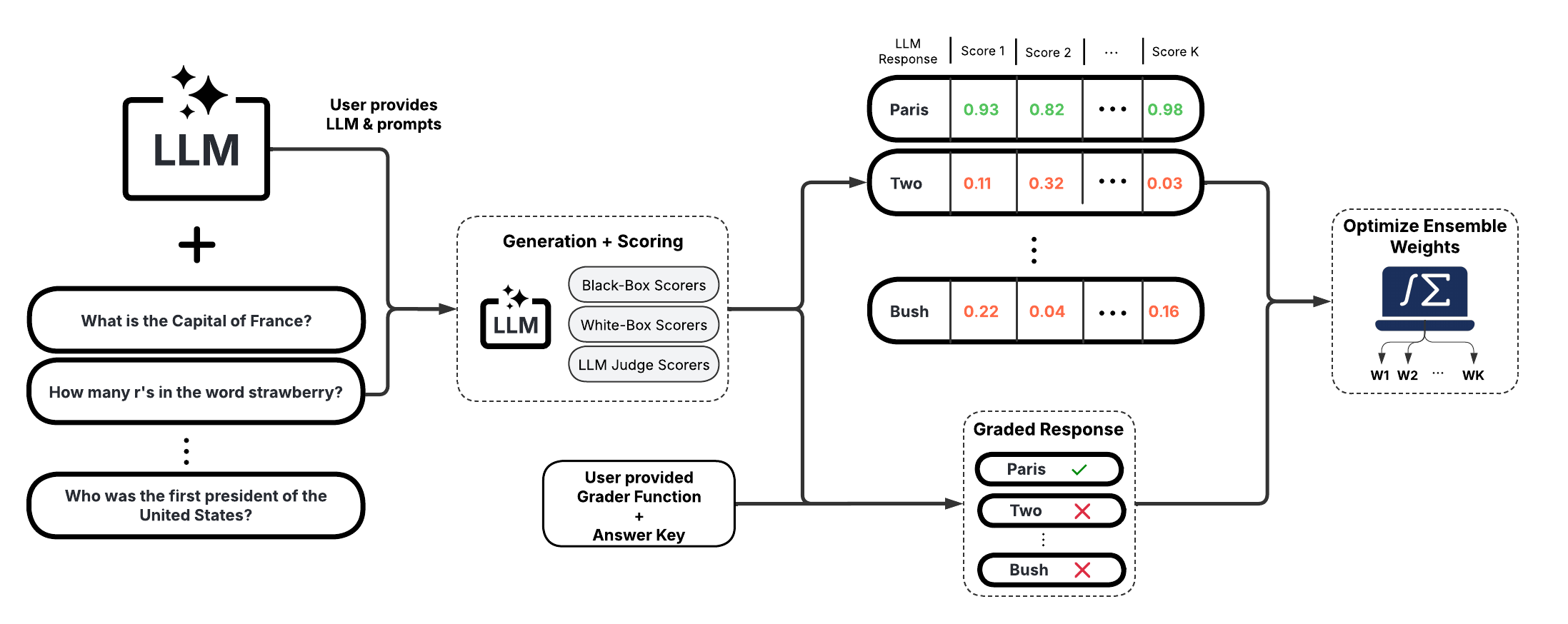}
    \caption{Illustration of Ensemble Tuning}
    \label{fig:ensemble_score}
\end{figure}

Once the binary grades (`correct' or `incorrect') are obtained, an optimization routine solves for the optimal weights according to a specified classification objective. The objective function may be threshold-agnostic, such as ROC-AUC, or threshold-dependent, such as F1-score. After completing the optimization routine, the optimized weights are stored as class attributes to be used for subsequent scoring. Below is a minimal example illustrating this process.

\begin{lstlisting}[language=Python]
from uqlm import UQEnsemble
## ---Option 1: Off-the-Shelf Ensemble (Chen & Mueller, 2023)---
# uqe = UQEnsemble(llm=llm)
# results = await uqe.generate_and_score(prompts=prompts, num_responses=5)

## ---Option 2: Tuned Ensemble---
scorers = [ # specify which scorers to include
    "exact_match", "noncontradiction", # black-box scorers
    "min_probability", # white-box scorer
    llm # use same LLM as a judge
]
uqe = UQEnsemble(llm=llm, scorers=scorers)

# Tune on tuning prompts with provided ground truth answers
tune_results = await uqe.tune(
    prompts=tuning_prompts, ground_truth_answers=ground_truth_answers
)
# ensemble is now tuned - generate responses on new prompts
results = await uqe.generate_and_score(prompts=prompts)
results.to_df()
\end{lstlisting}

\section{Compatible Models}
\label{sec:models}

UQLM integrates with the LangChain Chat Model interface, making it compatible with a large and evolving ecosystem of providers and models; an exhaustive list is infeasible. In practice, users access models through LangChain wrappers for commercial API providers such as OpenAI, Anthropic, Google, Cohere, and AWS Bedrock, as well as for locally deployed open-source models like Llama 3, Mistral, and Qwen using tools such as Ollama. For an up-to-date catalog and provider-specific options, we refer readers to the LangChain documentation for supported chat models.\footnote{\url{https://python.langchain.com/docs/integrations/chat/}} Note that white-box uncertainty methods require access to token log probabilities, which some wrappers do not expose. LangChain documents how to request these log probabilities and how to verify whether a specific wrapper returns them.\footnote{\url{https://python.langchain.com/docs/how_to/logprobs/}} Table \ref{tab: compatibility} provides detailed compatibility information for each scorer family.

\begin{table}[h]
\centering
\label{tab: compatibility}
\label{tab:uq_method_requirements}
\begin{tabular}{lcp{6.3cm}}
\toprule
\textbf{Scorer family} & \textbf{Needs logprobs} & \textbf{Compatibility} \\
\midrule
Black-Box UQ & No & Compatible with all chat models \\
White-Box UQ & Yes & Chat models that expose logprobs \\
LLM-as-a-Judge & No & Compatible with all chat models \\
UQ Ensemble & Depends & Depends on included scorers \\
\bottomrule
\end{tabular}
\caption{UQLM method families and interface requirements.}
\end{table}

\newpage

\bibliography{ref}

\end{document}